\pdfoutput=1

\documentclass[11pt]{article}

\usepackage{EMNLP2023}

\usepackage{times}
\usepackage{latexsym}
\usepackage{physics}
\usepackage{breqn}
\usepackage{tablefootnote}

\usepackage[T1]{fontenc}

\usepackage[utf8]{inputenc}

\usepackage{microtype}

\usepackage{inconsolata}

\usepackage{xspace}

\usepackage{array,multirow,graphicx}

\definecolor{amber}{rgb}{1.0,0.49,0.0}

\newcommand{\STAB}[1]{\begin{tabular}{@{}c@{}}#1\end{tabular}}

\newcommand\cnn{CNN/DailyMail\xspace}
\newcommand\xsum{XSUM\xspace}
\newcommand\reddit{RedditTIFU-long\xspace}
\newcommand\samsum{SAMSum\xspace}
\newcommand\forumsum{ForumSum\xspace}

\usepackage{amsmath,amsfonts,bm}

\def\eqref#1{equation~\ref{#1}}

\def\1{\bm{1}}

\DeclareMathAlphabet{\mathsfit}{\encodingdefault}{\sfdefault}{m}{sl}
\SetMathAlphabet{\mathsfit}{bold}{\encodingdefault}{\sfdefault}{bx}{n}

\newcommand\seqx{\mathbf{x}}

\newcommand\seqyt{\bar{\mathbf{y}}}

\newcommand\seqyg{\hat{\mathbf{y}}}
\newcommand\seqeg{\hat{\mathbf{e}}}

\newcommand\lcal{L^{\mathrm{cal}}}
\newcommand\lreg{L^{\mathrm{reg}}}

\title{Calibrating Likelihoods towards Consistency in Summarization Models}

\author{ Polina Zablotskaia\thanks{~~Equal contribution.} \quad Misha Khalman\footnotemark[1] \quad   Rishabh Joshi  \quad Livio Baldini Soares \\ \quad \bf{Shoshana Jakobovits}  {\bf\quad Joshua Maynez\quad Shashi Narayan} \quad {\bf } \\
Google DeepMind\\
\texttt{\small{\{polinaz,khalman,rishabhjoshi,liviobs,jakobovits,joshuahm,shashinarayan\}@google.com}}
}
\begin{document}
\maketitle  
\begin{abstract}


Despite the recent advances in abstractive text summarization, 
current summarization models still suffer from generating factually inconsistent summaries, reducing their utility for real-world application. We argue that the main reason for such behavior is that the summarization models trained with maximum likelihood objective
do not accurately rank sequences by their consistency. 
In this work, we 
solve this problem by calibrating the likelihood of model generated sequences to better align with a consistency metric measured by natural language inference (NLI) models. 
The human evaluation study and automatic metrics show that the calibrated models generate more consistent and higher-quality summaries.
We also show that the models trained using our method return probabilities that are better aligned with the NLI scores, which significantly increase reliability of summarization models.

\end{abstract}

\section{Introduction}

Recent years have witnessed a huge leap forward in abstractive summarization \cite{pegasus,liu-etal-2022-brio}, yet the wider adaptation of summarization models is limited by their tendency to generate {\em hallucinations} – outputs with contradicting or unsupported information to their input article \cite{falke-etal-2019-ranking,maynez-etal-2020-faithfulness}. Hallucinations in summarization models can be mostly attributed to two main reasons. First, most summarization systems are trained to maximize the log-likelihood of the reference summary, which does not necessarily reward models for being faithful. Moreover, models are usually agnostic to the noises or artifacts of the training data, such as reference divergence, making them vulnerable to hallucinations \cite{kryscinski-etal-2019-neural,dhingra-etal-2019-handling}. Thus, models can generate texts that are not consistent with the input, yet would likely have reasonable model log-likelihood. We refer to this phenomenon as models' sequence likelihood not being calibrated to their consistency.

\begin{figure}[t!]
    \centering
    \footnotesize
    \resizebox{0.99\columnwidth}{!}{
    \begin{tabular}{@{}p{8cm}@{}} \hline
    \textbf{Input:} The man, from Aberdeen, was charged after ``suspicious incidents'' in the Aberdeen, Aberdeenshire and Montrose areas. A report has been submitted to the procurator fiscal. Sgt Andy Peerless, of Police Scotland, said: ``The information provided to us from the public was vital.'' \\
    \textbf{Before:} A \textcolor{amber}{49-year-old} man has been charged with a number of offences including \textcolor{amber}{rape and possession of a bladed article}. \\
    \textbf{After:} A man has been charged following ``suspicious incidents'' in Aberdeen, Aberdeenshire and Montrose. \\  
    \hline 
    \hline
    \textbf{Input:} Violet D'Mello entered the enclosure for a photo next to the cats at the Kragga Kamma Game Park in Port Elizabeth earlier this year. She suffered injuries to her head, stomach and legs during the incident. The authorities in South Africa have ruled the park was not negligent. A party of visiting schoolboys and a cheetah in heat were said to have been factors in what happened. Mrs D'Mello, 60, said she survived by ``playing dead''. She had been on holiday with her husband Archie at the time. \\	
    \textbf{Before:} A woman who was \textcolor{amber}{mauled to death} by a cheetah at a South African game park has been \textcolor{amber}{awarded a six-figure sum of money.} \\
    \textbf{After:}  A woman who suffered injuries when she entered a cheetah enclosure at a South African game park has said she "played dead". \\ 
    \hline
    \end{tabular}
    } 
    \caption{XSum inputs and system generated summaries before and after calibration. The text spans in \textcolor{amber}{amber} are hallucinated.}
    \label{fig:xsum_examples} 
    \vspace{-0.3cm}
\end{figure}

The textual entailment score -- the entailment probability of a summary (hypothesis) given its input (premise) -- has been widely used to quantify the extent to which generated summaries are faithful or consistent to their input \cite{falke-etal-2019-ranking,maynez-etal-2020-faithfulness,narayan-etal-2022-well,honovich-etal-2022-true-evaluating}.
Unsurprisingly, several efforts aiming to calibrate summarization models towards consistency focus on textual entailment signals. \newcite{pasunuru-etal-2017-towards} use the {\em multi-task learning} to jointly train their decoder as a summary generator as well as an entailment classifier. 
\newcite{pasunuru-bansal-2018-multi} proposed to use {\em reinforcement learning} with sequence-level reward for entailment optimizing models to assign higher probability to logically-entailed summaries. 
{\em Reranking-based approach} \cite{falke-etal-2019-ranking} uses a two-stage reranking system that first generates candidate summaries and then uses textual entailment predictions to detect consistency errors and rerank alternative predicted summaries.
Another trend proposes to leverage consistency signals via {\em controlled generation} \cite{kesker_2019_ctrl,rashkin-etal-2021-increasing} to calibrate summarization models. Specifically, training examples are supplemented by prepending special tokens to inputs to indicate/control whether the output should be entailed or not. This way model is better calibrated in differentiating inconsistent examples from consistent examples. 
Some have also relied on {\em data filtering} where we only train on examples whose summaries are predicted to be entailed by the input \cite{narayan-etal-2021-planning,aharoni2022mface}.

Recently \newcite{liu-etal-2022-brio} introduced calibration methods to align candidates' sequence likelihoods to their quality as measured by their similarities to the target sequence. First they decode candidates from a fine-tuned model on its own training dataset, and then continue training the model with a multi-task learning objective of sequence candidates with contrastive reranking  and  token-level  generation. \newcite{liu-etal-2022-brio} used metrics like Rouge \cite{lin-2004-rouge} and BERTScore \cite{bertscore} to rank different decoded candidates with their similarities to the target sequence. \newcite{slic_iclr23} generalizes \newcite{liu-etal-2022-brio} and uses their similarities to the target sequence in the model's latent space, instead of relying on external metrics like Rouge and BERTScore. Both \newcite{slic_iclr23} and \newcite{liu-etal-2022-brio} demonstrate that their methods significantly improve the quality of generated summaries when evaluated against target sequences using Rouge or BERTScore. However, the improvements in the similarities to the target sequence doesn't necessarily lead to consistent summaries. Figure~\ref{fig:xsum_examples} presents few hallucinated (spans mark in \textcolor{amber}{amber}) summaries generated using these methods. 

In this paper we propose {\em \textbf{S}equence {\textbf{Li}kelihood \textbf{C}alibration with \textbf{NLI}} (or \textbf{SLiC-NLI})} to calibrate summaries' sequence likelihood to their consistency. Our approach builds on \cite{slic_iclr23} and \cite{liu-etal-2022-brio} but uses textual entailment scores to rank candidate summaries, instead of Rouge or BERTScore. In particular, we 
decode candidates from a fine-tuned model on its own training dataset, estimate entailment probabilities of candidate summaries given their respective inputs, and  then continue training the model with a multi-task learning objective of sequence candidates with contrastive reranking and token-level generation. 

Unlike reinforcement learning, it is a one-time offline process that avoids costly online decoding processes. Also, when compared to two-stage reranking systems, it doesn't require a separate reranking model that incurs additional complexity and compute.

We experimented with five different abstractive summarization tasks: CNN/DailyMail \cite{hermann2015cnndm}, ForumSum \cite{khalman-etal-2021-forumsum-multi}, RedditTIFU-long \cite{kim-etal-2019-abstractive}, SAMSum \cite{gliwa-etal-2019-samsum} and XSUM \cite{narayan-etal-2018-dont}, due to their diversity in domain, style, abstractiveness, and summary lengths. We show that using our approach models can generate better consistent summaries,  without sacrificing their overall quality when evaluated automatically and by humans. 


\section{Related Work}

\subsection{Measuring Consistency}

A large number of approaches have been proposed for automatic detection of factual inconsistencies. Most notably, Natural Language Inference \cite{bowman2015large} approaches has been shown to have a large correlation with human consistency ratings on generation tasks, including summarization \cite{maynez-etal-2020-faithfulness, laban-etal-2022-summac, goyal-etal-2021-multi, goyal-durrett-2021-annotating}. Other approaches based on question generation and answering have been also shown to perform well in detecting factual consistency \cite{scialom-etal-2021-questeval, honovich-etal-2021-q2, deutsch-etal-2021-towards}, but usually require a pipeline of model inferences that makes them impractical for some applications.
Many studies have investigated automatic detection of factual inconsistencies in a wider variety of tasks \cite{honovich-etal-2022-true-evaluating, Tang2022UnderstandingFE}, and show that large-scale NLI models have among the highest agreement with human ratings. 

\subsection{Calibrating Consistency}

Model calibration is commonly used in classification tasks, whereas in sequence generation it has not being well defined generally. In our context, model calibration refers to aligning the sequence likelihood to the target entailment probability. 

\paragraph{Reranking-based Approach}

Many works have proposed to reranking as approach to Many works have proposed approaches that first decode a number of outputs and re-rank them as a second stage. \newcite{liu-liu-2021-simcls} decode outputs with diverse beam search and using a RoBERTa-based model to rank them next. Similarly in the neural machine translation (NMT), \newcite{fernandes-etal-2022-quality} and \newcite{lee-etal-2021-discriminative} train rerankers that mimic automatic metrics (BLEU, COMET and BLEURT) and re-rank top-k decodes accordingly. SummaReRanker \cite{ravaut-etal-2022-summareranker} found that performance is improved by training generation and reranking models on exclusive halves of the training data instead of on the same data.
BRIO \cite{liu-etal-2022-brio} includes sequence-to-sequence generation models for both generation and reranking stages. They rank different  candidates by their similarities to the target sequence using automatic metrics. \newcite{slic_iclr23} generalizes this idea by computing the similarities to the target sequence in the model's latent space. \newcite{Zhao2023SLiCHFSL} 

\paragraph{RL-based Approach}

Reinforcement learning has been proposed as an approach to optimize signals directly. \newcite{paulus2018a} optimize the evaluation metric ROUGE via RL fine-tuning. The authors found that optimizing for single discrete evaluation metric such as ROUGE can be detrimental to the model quality and fluency. \newcite{ziegler2019fine} and  \newcite{Stiennon2020LearningTS} trained reward models to learn human preference based on collected human judgments of competent fine-tuned models. Using PPO, the supervised policy is fine-tuned against the learned reward model. 
The authors found that this approach leads to better quality summaries than optimizing with respect to ROUGE.


\paragraph{Controllable Generation}

Controllable generation has been proposed as an approach to increase consistency. \newcite{he-yiu-2022-controllable} proposed the use of control codes to influence generated outputs to match desired characteristics such as style and length as they were observed in the training data. \newcite{rashkin-etal-2021-increasing} and \newcite{aharoni2022mface} extended this approach to increase consistency in grounded dialog and multilingual summarization, correspondingly, by adding a control feature based on inferred NLI scores given the summary and input document \cite{honovich-etal-2022-true-evaluating}.

\paragraph{Summary Generation with Planning}

\newcite{narayan-etal-2021-planning} proposed that intermediary plans, based on entities, are useful to increase grounding and consistency in summarization by avoiding common pitfalls seen in autoregressive generation. Moreover, sequence-to-sequence models can learn to produce those plans and the output summaries sequentially in an end-to-end manner. These plans are also controllable and models trained this way are able to produce summaries grounded to the modified plans. Further, \newcite{Narayan2022ConditionalGW} showed that plans based on questions and answers provide anchoring for more complex tasks, for instance multi-document summarization, aiding further on consistency of longer summaries.

\paragraph{Data Filtering Approach}

\newcite{narayan-etal-2021-planning} and \newcite{aharoni2022mface} additionally proposed a simple approach to filter the training data based on inferred NLI scores given the summary and input document. Using only a subset of the training data, inferred to be consistent with the input, model consistency by automatic metrics and human evaluations is improved.


\begin{figure}[t!]
\centering
    \includegraphics[width=\columnwidth]{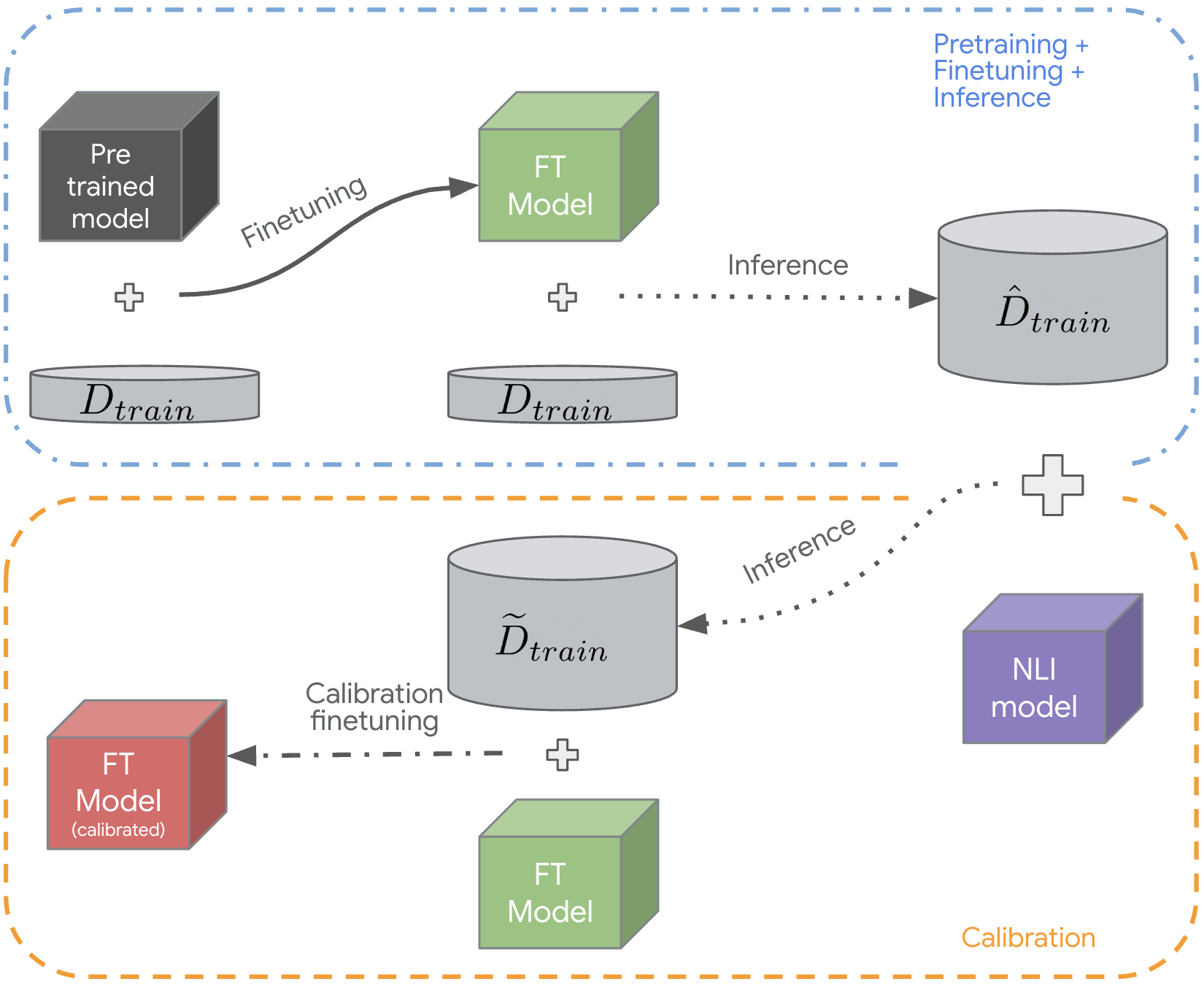}
\caption{Our method consists of two parts: \textbf{top} (blue color) represents the usual finetuning and inference and \textbf{bottom} (orange color) represents the SLIC-NLI methods consisting of the inference using the NLI model and the SLIC calibration.} 
\label{fig:diagram}

\end{figure}

\section{Method}

Following \newcite{slic_iclr23} and \newcite{liu-etal-2022-brio}, we introduce a third {\em calibration stage} to the popular paradigm of pretraining and fine-tuning, as explained in Figure~\ref{fig:diagram}.  Let $D_{train}: \{\seqx, \seqyt\}_n$ be the dataset used for fine-tuning. We first generate $m$ candidates $\{\seqyg\}_m$ for each training instance in $D_{train}$ from a fine-tuned model; we refer to this augmented dataset as $\hat{D}_{train}$ consisting of $\{\seqx, \{\seqyg\}_m, \seqyt \}_n$. We then calibrate the fine-tuned model by continuing training with the following loss:
\begin{align}
\mathcal{L}(\theta) = \sum_n & \lcal(\theta; \seqx, \{\seqyg\}_m, \seqyt) \nonumber \\
& + \lambda \lreg (\theta, \theta_{ft}; \seqx, \seqyt),
\end{align}
where $\theta$ and $\theta_{ft}$ are the current and finetuned model weights, $\lcal$ and $\lreg$ are the calibration and regularization losses, respectively. The calibration loss $\lcal$ aims to align models' decoded candidates' sequence likelihood $P_{\theta}(\seqyg | \seqx)$ according to their entailment scores, whereas the regularization loss $\lreg$ prevents models from deviating significantly from their fine-tuned model parameters.

\begin{figure*}[th!]
\includegraphics[width=16cm]{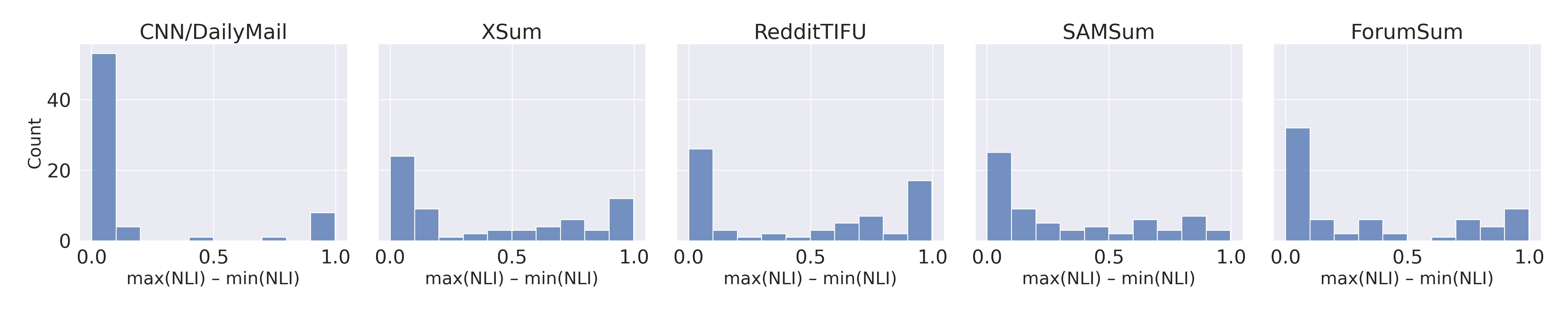}
\centering
\caption{Distribution of the NLI scores over the inference outputs with beam size $=15$. All dataset except for CNN/DailyMail have a diverse variety of generated summaries per document.}
\label{fig:score_dist}
\end{figure*}

\subsection{Calibrating towards Consistency}
\label{subsec:calnli}

In order to calibrate the model towards the consistency we annotate $\{\seqyg\}_m$ with textual entailment scores (Natural Language Inference or NLI) \citep{bowman-etal-2015-large}, i.e. we estimate entailment probabilities of candidate summaries given their respective inputs. To esimate the entailment we follow \citet{honovich-etal-2022-true-evaluating} and trained an NLI model by fine-tuning T5-11B \cite{t5} on the Adversarial NLI (ANLI; \citeauthor{nie-etal-2020-adversarial}, \citeyear{nie-etal-2020-adversarial}) dataset. In Figure~\ref{fig:diagram} the dataset $\hat{D}_{train}$ annotated with entailment probabilities is represented as $\tilde{D}_{train}: \{\seqx, \{\seqyg, \seqeg\}_m, \seqyt \}_n$, where $\seqeg$ is the entailment score of the candidate $\seqyg$. 
Figure~\ref{fig:score_dist} shows how the NLI scores are distributed across different datasets, overall we found out that for every dataset except for CNN/DailyMail we have good representation of good and bad examples for effective calibration. The calibration loss 
\begin{align}
\lcal =  \max(0, \beta & - \log P_\theta(\seqyg_+ | \seqx) \nonumber \\ 
& + \log P_\theta(\seqyg_- | \seqx)) \label{eq:len_cal}
\end{align}
then trains the model to learn the ranking among candidates pairs $(\seqyg_+, \seqyg_-)$, uniformly sampled from $ \{\seqyg\}_m$, according to their entailment scores. In this case, $\seqyg_+$ ranks highers than  $\seqyg_-$ as $\seqeg_+ > \seqeg_-$. 

Our approach differs from 
\newcite{slic_iclr23} and \newcite{liu-etal-2022-brio} where they proposed to use
the similarity between the candidate $\seqyg$ and the target $\seqyt$ conditioned on the context $\seqx$ to get ranking among candidate pairs, instead of textual entailment scores.


\subsection{Length regularization}
As a result of our extensive experimentation with the various $\beta$'s we have made curious observation about NLI scores. It appears that there is a slight positive correlation between the  length of the generated summaries and NLI. However this phenomena could be seen as a way of the model to "cheat" and over-optimize in the direction of the higher NLI. Oftentimes a dramatic increase in the length can come out of the repetition of the same sentences over and over. Naturally we would like to avoid this behavior. In pursuit of containing the length of the generating summaries we experiment with an additional length regularization term. We have experimentally found out that it is best to compare the length of the generated sequence $\seqyg$ with the length of the target sequence $\seqyt$ via simple ratio: 
\begin{dmath}
    f_{len}(\seqyg) = \left( 1 - \left|1 - \frac{l(\seqyg)}{l(\seqyt)} \right| \right), 
\end{dmath}
where $l(y)$ is the length of the sequence $y$. We subsequently update our calibration loss $\lcal$ from (\ref{eq:len_cal}) using $f_{len}$ to scale the log-likelihoods, up-weighted with  $\alpha$:
\begin{align}
\lcal =  & \max(0, \beta - \alpha \cdot f_{len}(\seqyg_+) \cdot \log P_\theta(\seqyg_+ | \seqx) \nonumber \\ 
& + \alpha \cdot  f_{len}(\seqyg_-) \cdot\log P_\theta(\seqyg_- | \seqx))  \label{eq:len_cal_reg}
\end{align}
Finally, for the regularization loss $\lreg$ we follow \newcite{slic_iclr23} and use the KL divergence loss minimizing the probability distribution distance between the calibrated model and the fine-tuned model at each token on the target sequence. \newcite{liu-etal-2022-brio} proposed to use the cross-entropy loss as the regularization loss. Nevertheless, both losses have been shown to perform similarly for summarization \cite{slic_iclr23}. 


\section{Experimental Setup}

\subsection{Summarization Datasets}

We have experimented with a diverse set of summarization datasets, with respect to different domains, styles, abstractivenesses, and summary lengths.

\paragraph{\textbf{\cnn}} \cite{hermann2015cnndm,see-etal-2017-get} summarization dataset contains 313k articles from the CNN and Daily Mail newspapers with bullet point summaries. The summaries are on average 3-4 sentences and relatively extractive.

 \paragraph{\textbf{\forumsum}} \cite{khalman-etal-2021-forumsum-multi} summarization dataset contains 4058 conversations from a wide variety of internet forums and their high-quality human written summaries. 

\paragraph{\textbf{\reddit}} \cite{kim-etal-2019-reddittifu} summarization dataset contains 42k posts of informal stories from sub-reddit TIFU from 2013-Jan to 2018-Mar with author written summaries. The style and length of the summaries are very diverse.

\paragraph{\textbf{\samsum}} \cite{gliwa-etal-2019-samsum} summarization dataset contains 16k high-quality chat-dialogues and their summaries written by linguists.

\paragraph{\textbf{\xsum}} \cite{narayan-etal-2018-dont} summarization dataset consists of 227k
BBC articles from 2010 to 2017 with a single sentence highly abstractive summary. Sometimes the summary contains information not present in the article.







\subsection{Automatic Evaluation}

We report on ROUGE \cite{lin-2004-rouge} which is commonly used to measure the informativeness and fluency of model generated summaries against gold-standard references. 
 
We report on the reference-free NLI score as a proxy for faithfulness \cite{maynez-etal-2020-faithfulness,honovich-etal-2022-true}. Regarding NLI, we compute for each summary whether it is entailed by the input, and report the average over all examples. We use the same NLI model 
that we use for calibration as described in \S\ref{subsec:calnli}.



\subsection{Human Evaluation}

We conducted human evaluation of the generated summaries for all 5 datasets. We picked our 3 models: finetuned, best calibrated (Eq~\ref{eq:len_cal} for $\lcal$) and best calibrated with length regularization (Eq~\ref{eq:len_cal_reg} for $\lcal$), along with other baselines. 
For each dataset we sampled 100 examples from its corresponding test set. For each example we generate summaries using different models and send to crowd-workers for side-by-side quality annotation. We present our raters a document and model generated summaries, and ask them to assess each summary individually for overall {\em quality} (on a scale of 1:Poor Summary  to 5:Great summary)) and {\em factuality} (a binary decision assessing whether everything in the summary can be verified in the document).
Each assessment is replicated by three different crowd workers. 
For quality we average the annotated scores across all replicas of each task. For the factuality metric we aggregate the metric using majority vote. The models are anonymized and randomly shuffled to avoid biases in the annotation.
For more details about the human evaluation template see Appendix~\ref{sec:appendix_human_eval}.

\subsection{Implementation Details}

We experimented with T5 (large, 500M parameters) with a maximum input sequence length of 1,024 tokens and a maximum output length of 256 tokens. We trained all our models with a leaning rate of 0.001 and a batch size of 128, for 50K steps. We select best checkpoints using average Rouge performance on validation sets, unless specified otherwise. During inference, we use beam search with size 5 and alpha 0.8.

\section{Results}

\paragraph{Ablation on Calibration Weights and its Effect on Lengths}

\begin{table*}[ht!]
    \centering
    \footnotesize
    {
    \begin{tabular}{cccccccc}
    \hline
     \textbf{Dataset} & \textbf{$\beta$} & \textbf{NLI \%}  & \textbf{NLI gain \%} & \textbf{R1/R2/RL} & \textbf{Coverage \% } &\textbf{Length} & \textbf{Repetition \%} \\ \hline
        \multirow{4}{*}{\STAB{\rotatebox[origin=c]{90}{ForumSum}}}  & $10^{-3}$  &  \textbf{82.13} & \textbf{ 10.47} & 38.51 / 18.08 / 31.16& 88.5 & 43.25 & 19.10
\\
         & $10^{-4}$ & 78.15 & 6.50 & \textbf{40.74 / 20.09 / 33.59}& 86.9  & 32.28 & 13.10\\ 
         & $10^{-5}$ & 75.05 & 3.39 & 40.50 / 19.39 / 32.97& 84.5 & 27.17 & 7.20
\\ 
         & w/o & 71.66 & 0.00  & 39.82 / 18.74 / 32.37& 83.6 & 25.03 & 6.40 \\  \hline
         \multirow{4}{*}{\STAB{\rotatebox[origin=c]{90}{RedditTifu}}}  &  $10^{-3}$ &  \textbf{89.43} & \textbf{ 8.23} & 27.28 / 8.65 / 21.60& 93.9 & 23.76 & 7.30 \\ 
         & $10^{-4}$ & 84.45 & 3.24  & 29.96 / 10.82 / 24.82 & 90.9 & 16.34 & 2.40 \\ 
         &  $10^{-5}$ & 82.28 & 1.07 & 30.02 / \textbf{10.85 / 25.05}& 89.2  & 15.27 & 1.30
 \\ 
         & w/o & 81.21 & 0.00 & \textbf{30.22} / 10.70 / 24.63 & 88.9 & 16.28 & 1.80 \\  \hline
        \multirow{4}{*}{\STAB{\rotatebox[origin=c]{90}{SAMSum}}} &  $10^{-2}$ &  \textbf{96.14} &  \textbf{9.41}& 48.93 / 24.68 / 39.76 & 81.3 & 29.08 & 4.10
\\ 
         &  $3 \cdot 10^{-4}$ & 91.51 & 4.78 & \textbf{54.47 / 30.15} / 45.72 & 80.4 & 19.74& 1.60 \\ 
        & $10^{-4}$& 87.93 & 1.19 & 54.33 / 29.98 / \textbf{45.85}& 79.6  & 17.92 & 1.50
\\ 
        & w/o & 86.73 & 0.00 & 54.52 / 30.09 / 45.75& 79.2  & 18.93 & 1.70\\  \hline
        \multirow{4}{*}{\STAB{\rotatebox[origin=c]{90}{XSUM}}}  & $10^{-2}$&  \textbf{81.21} &  \textbf{28.19} & 39.46 / 16.92 / 31.88 & 83.0 & 18.07 & 0.40
\\ 
        & $3 \cdot 10^{-4}$ & 77.46 & 24.44  & 41.32 /18.77 / 33.71 & 80.9 & 17.33 & 0.40\\ 
        &  $10^{-3}$ & 57.21 & 4.19 & \textbf{44.80 / 21.93 / 36.99} & 74.1 & 17.22 & 0.40\\ 
        & w/o & 53.02 & 0.00 & 44.73 / 21.88 / 36.94 & 73.4 & 16.94 & 0.40
 \\  \hline
        \multirow{4}{*}{\STAB{\rotatebox[origin=c]{90}{\parbox{1.3cm}{\centering CNN/\\DailyMail}}}} &  $10^{-2}$&  \textbf{89.47} &  \textbf{2.12}& 42.41 / 20.25 / 29.78 & 99.4 & 68.68 & 18.50 \\ 
        &  $10^{-3}$ & 89.08 & 1.72 & 42.96 / 20.79 / 30.28 & 99.4 & 70.42 & 17.60
 \\ 
        & $3 \cdot 10^{-4}$ & 88.57 & 1.22 & 43.52 / 21.23 / 30.78 & 99.3 & 69.26 & 14.10
\\ 
        & w/o & 87.36 & 0.0 & \textbf{44.29 / 21.82 / 31.62 }& 99.2 & 57.13 & 3.90\\ \hline
    \end{tabular}
    }
    \caption{
    The effect of different calibration weights on the model performance in terms of NLI. We also report on other automatic measures: Rouge-1, Rouge-2 and Rouge-L scores (R1/R2/RL), Coverage (percentage of tokens in the generated  summary that appeared in the input), Repetition (percentage of repeated tokens in the output summary) and the summary lengths. All the results are reported on respective validation sets.
    }
    \label{tab:nli_main}
    \vspace{-0.3cm}
\end{table*}

\begin{figure*}[t!]
\includegraphics[width=16cm]{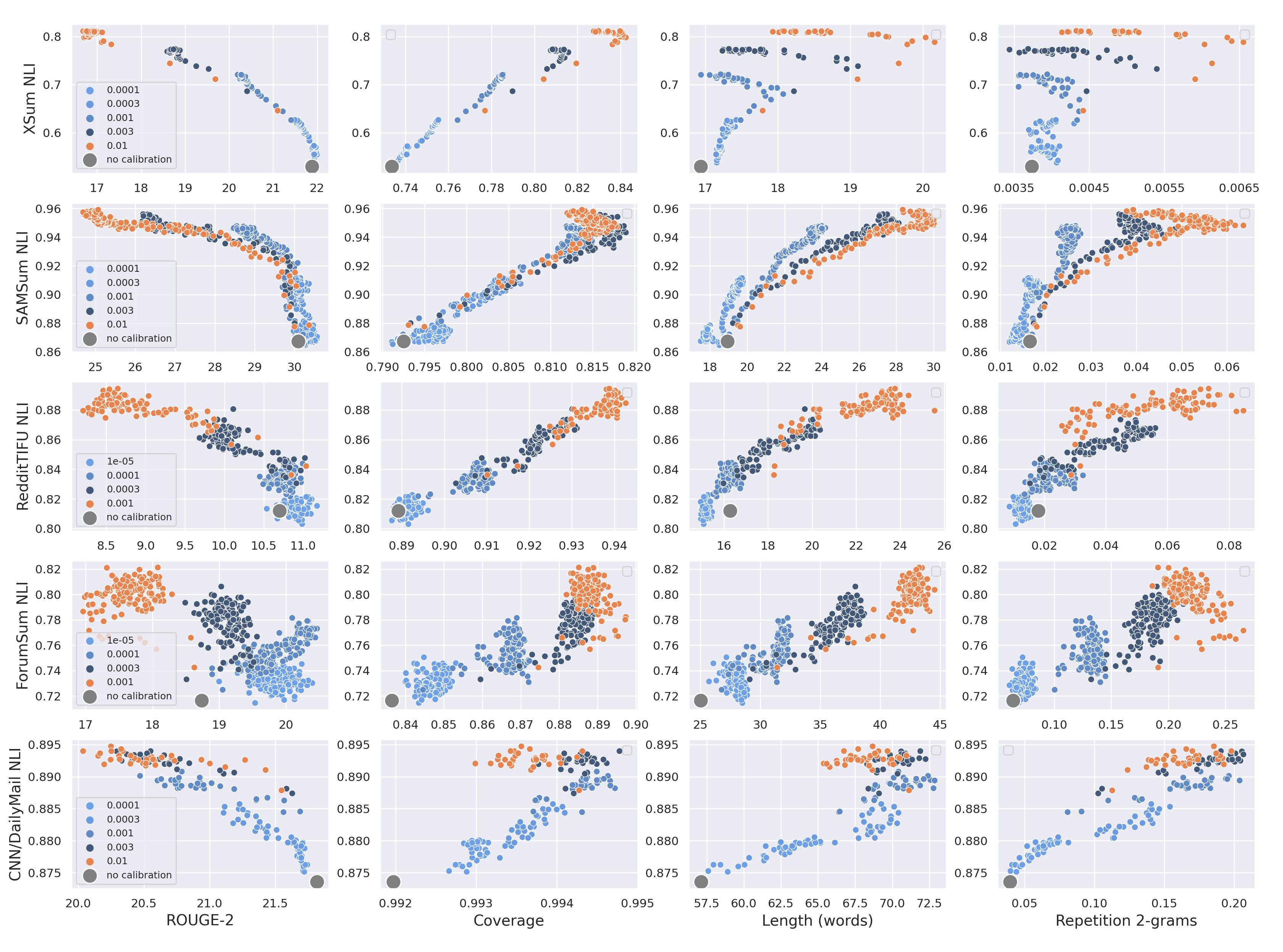}
\centering
\caption{Pareto frontier that demonstrates the trade-offs between NLI and other metrics such as Rouge-2, Coverage, Length and Repetition on all datasets. \label{fig:pareto_selected}}
\end{figure*}


\begin{table*}[t!]
    \centering
    \footnotesize
    {
    \begin{tabular}{ccccccc}
    \hline
        \textbf{$\alpha$} &\textbf{NLI \%}  & \textbf{NLI gain \%} & \textbf{R1/R2/RL} & \textbf{Coverage \%} & 
        \textbf{Length} & \textbf{Avg} \\ \hline
        100 & 66.30 & 13.28 & 42.76/19.73/34.44 & 77.40 & 20.02 & 0.24 \\ 
        10 & 68.24 & 15.22 & 42.73/19.68/34.44 & 77.80 & 19.62 & 0.32 \\ 
        1 & 78.34 & 25.3 & 40.72/17.80/32.87 & 82.60 & 18.54 & 0.62 \\ 
        \textbf{0.5} & \textbf{78.87} & \textbf{25.85} & \textbf{40.17/17.64/32.80} & \textbf{82.20} & \textbf{16.82 }& \textbf{0.81} \\
        0.1 & 74.28 & 21.27 & 41.36/19.22/34.11 & 79.80 & 15.69 & 0.73 \\ 
        0.01 & 56.35 & 3.33 & 44.82/21.96/37.02 & 74.00 & 17.02 & 0.41 \\ 
        $10^{-3}$ & 53.62 & 0.60  & 44.89/21.99/37.06 & 73.30 & 17.16 & 0.34\\ 
        $10^{-4}$ & 53.39 & 0.37 & 44.86/21.93/37.00 & 73.30 & 17.21 & 0.33 \\\hline
        w/o $f_{len}$ & 81.21 & 28.19 & 39.46/16.92/31.88 & 83.00 & 18.07 & 0.73 \\ \hline 
        w/o $\lcal$  & 53.02 & 0.00  & 44.73/21.88/36.94 & 73.40 & 16.94 & 0.36 \\ \hline
    \end{tabular}
    }
     \caption{
     The effect of various length regularizer weights on the XSum dataset performance. 
     We choose  $\beta = 0.5$ with the highest NLI scores of 78.87\% on the XSum validation set.} 
     \label{tab:len_cal_reg}
     \vspace{-0.13in}
\end{table*}

\begin{figure*}[t!]
\centering
\begin{tabular}{ccc}
\includegraphics[width=5cm]{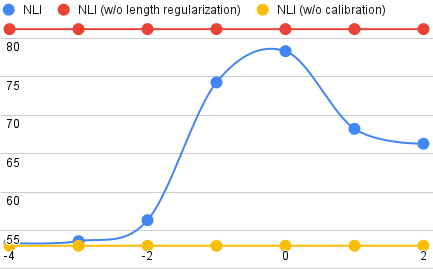} &
\includegraphics[width=5cm]{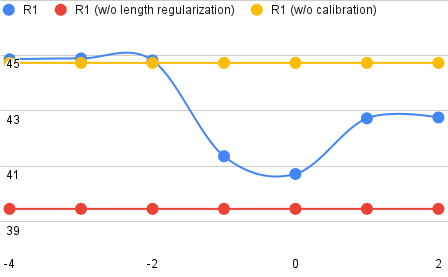} &
\includegraphics[width=5cm]{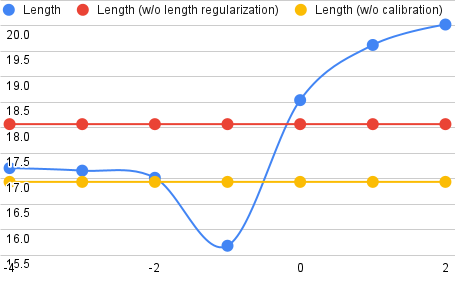}
\end{tabular}
\caption{Plot of NLI, Rouge (R1) and Summary lengths with 
various length regularizer weights. See Table~\ref{tab:len_cal_reg} for exact numbers.  \label{fig:plot_length_regularizer}}

\end{figure*}

We first ablate the effect of different calibration weights ($\beta$) in Eq~\ref{eq:len_cal} without applying the length regularization. Table~\ref{tab:nli_main} presents our results.

We achieve up to $\approx  30 \%$ improvement in terms of the NLI scores on XSUM datasets, $10.47 \%$ on \forumsum, $9.41 \%$ on \samsum, $8.23 \%$ on \reddit, and, $2.12 \%$ on \cnn.
Using different values of $\beta$ in $L^{cal}$ allows to control the level of the calibration, i.e. the bold colors in  Table~\ref{tab:nli_main} always correspond to the highest weight. We observe that higher calibration can often times affect the other metrics, for example ROUGE scores slightly decrease with the intensity of calibration, which can be non-desirable. 
Similar phenomenon can be seen with the increase in length and repetition which are can be a symptom of the model trying to "cheat" the NLI metric.  On Figure~\ref{fig:pareto_selected} we demonstrate the Pareto frontier that allows us to explore the optimal tradeoff between the NLI scores and various metrics.

\paragraph{Ablation on Length Regularizer}

In order to prevent the model from overfitting to the NLI metric we conduct an extensive set of experiments to analyze the effect of the length regularization on the various metrics.  As per Eq.~\ref{eq:len_cal_reg} we choose various $\alpha$ in order to increase the effect of regularization. Results are presented in Table~\ref{tab:len_cal_reg} and Figure~\ref{fig:plot_length_regularizer}. When $\alpha$ is very small, the model performs similar to its uncalibrated counterpart. But as we increase $\alpha$, we start seeing the effect of joint consistency and length calibration.
In order to pick the best configuration that equally opimizes for NLI and does not deviate much on length we propose an average score $Avg_{\alpha} = \frac{\textrm{NLI}'_{\alpha} + (1-\max(\textrm{L}'_{\alpha}, \textrm{L}'_{\textrm{w/o}})}{2}$, where $X'_{\alpha}= \frac{X_{\alpha} - \min(\mathbf{X})}{\max(\mathbf{X}) - \min(\mathbf{X})}$, i.e. simple min-max normalization. $\mathbf{X}$ is the set of values with different values of $\alpha$. Table~\ref{tab:len_cal_reg} highlights best scores that indicate the best results according to this metric. We follow the same recipe to select the best models for all datasets.




\begin{table*}[t!]
    \centering
    \footnotesize
    {
    \begin{tabular}{ccccc}
    \hline
     \textbf{Models} & \textbf{NLI \%}  & \textbf{R1/R2/RL}  &\textbf{Length} & \textbf{Repetition \%} \\ \hline
     \multicolumn{5}{c}{\textbf{\xsum}} \\ \hline 
     Pegasus                        &  54.00 &  46.23 / 24.21 / 38.64 &  18.91 &       0.45 \\
            Brio \footnotemark 
            &  49.76 &  47.22 / 24.68 / 39.28 &  19.42 &       0.81 \\
            SLiC  & 51.93 & 43.96 / 20.80 / 35.99 \footnotemark  & 16.85 & 0.50\\
            Cliff                          &  56.11 &  43.10 / 20.90 / 35.61 &  18.27 &       0.31 \\
            FactPegasus                    &  52.37 &  37.13 / 15.08 / 30.33 &  16.57 &       0.32 \\
            Frost (Drop)                &  58.75 &  43.58 / 20.94 / 36.39 &  17.51 &       0.30 \\ \hline
            Finetuned (w/o $\lcal$)        &  48.52 &  44.02 / 22.07 / 36.64 &  17.75 &       0.35 \\
            SLiC-NLI (w/o $f_{len}$)       &  \textbf{80.01} &  38.16 / 16.43 / 30.97 &  18.59 &       0.59 \\
            SLiC-NLI (with $f_{len}$).     &  74.16 &  40.10 / 18.86 / 33.34 &  15.74 &       0.32 \\
            \hline
        \multicolumn{5}{c}{\textbf{\cnn}} \\ \hline 
            Pegasus                        &  93.31 &  42.22 / 21.06 / 39.45 &  61.50 &       3.48 \\
            Brio                           &  88.75 &  46.30 / 23.25 / 31.93 &  63.09 &       3.27 \\
            SLiC & 93.38 & 43.86 / 21.18 / 30.88 & 52.59 & 3.80 \\
            Cliff                          &  91.08 &  33.91 / 14.29 / 24.27 &  51.43 &       1.45 \\
            Frost (Drop)                &  93.49 &  43.50 / 21.56 / 40.83 &  57.54 &       3.46 \\
            \hline
            Finetuned (w/o $\lcal$)  & 92.61 & 42.39 / 20.90 / 35.29 &  51.13 &       2.70 \\
            SLiC-NLI (w/o $f_{len}$)  &  \textbf{94.58} &  40.84 / 19.54 / 38.30 &  66.48 &      15.68 \\
            SLiC-NLI (with $f_{len}$) &  94.22 &  41.62 / 19.83 / 38.55 &  63.63 &       6.12 \\
        \hline 
        \multicolumn{5}{c}{\textbf{\forumsum}} \\ \hline
            SLiC & 75.78 & 41.44 / 20.08 / 34.22 & 35.85 & 14.03\\
            Finetuned (w/o $\lcal$)  &  72.97 & 40.34 / 19.29 / 32.66 &  31.66 &       7.88 \\
            SLiC-NLI (w/o $f_{len}$)  &  \textbf{78.26} &  38.74 / 19.15 / 32.25 &  38.12 &      18.47 \\
            SLiC-NLI (with $f_{len}$) &  77.82 &  40.82 / 20.22 / 34.07 &  30.61 &       8.59 \\
         \hline 
        \multicolumn{5}{c}{\textbf{\samsum}} \\ \hline 
            SLiC & 73.25 & 52.82 / 27.96 / 43.81 & 17.84 & 1.89\\
            Finetuned (w/o $\lcal$)   &  73.00 &  51.01 / 26.36 / 42.54 &  18.62 &       2.04 \\
            SLiC-NLI (w/o $f_{len}$)  &  \textbf{86.57} &  46.44 / 22.60 / 37.66 &  30.00 &       5.28 \\
            SLiC-NLI (with $f_{len}$) &  84.63 &  49.79 / 25.39 / 41.51 &  20.59 &       1.65 \\
        \hline 
        
        \multicolumn{5}{c}{\textbf{\reddit}} \\ \hline
            SLiC & 75.61 & 27.51 / 7.98 / 21.71 & 16.22 & 1.20\\
            Finetuned (w/o $\lcal$)   &  69.10 &   27.52 / 9.16 / 22.53 &  14.54 &       0.70 \\
            SLiC-NLI (w/o $f_{len}$)  &  \textbf{85.75} &   27.40 / 9.01 / 22.33 &  18.92 &       3.81 \\
            SLiC-NLI (with $f_{len}$) &  80.87 &   27.43 / 9.35 / 22.78 &  15.40 &       2.15 \\
        \hline 
    \end{tabular}
    }
    \caption{Final results on various test sets. 
    We include results from several state-of-the-art summarization models such as Pegasus \cite{pegasus}, Brio \cite{liu-etal-2022-brio},  SLiC \cite{slic_iclr23}, Cliff \cite{cao-wang-2021-cliff}, FactPegasus \cite{wan-bansal-2022-factpegasus} and Frost \cite{narayan-etal-2021-planning}. Cliff, FactPegasus and Frost are particularly trained or designed to generate factual summaries. For Frost, we report on Frost (Drop) which avoids hallucinated entities in summaries by dropping them form their entity plans. 
    In each dataset we consistently show outstanding results on NLI. Having shorter sequence can be motivated by the generation latency or the risk of repetition in the summaries, in that case SLiC-NLI variant with length regularisation can be used and it surpasses other baselines as well.}
    \label{tab:final_results}
\end{table*}

\begin{figure}[t!]
    \centering
    \footnotesize
    \resizebox{0.99\columnwidth}{!}{
    \begin{tabular}{@{}p{8cm}@{}} \hline
    \textbf{Input:} Police were alerted to the stabbing in Harehills Lane, Harehills, at about 15:40 GMT. The wounded teenager was taken to hospital for treatment, but died a short time later. A 15-year-old boy has been arrested on suspicion of murder, West Yorkshire Police said. He remains in custody for questioning. Det Supt Pat Twiggs, of West Yorkshire Police, said: This tragic incident happened in a busy area at a busy time of day with large numbers of people going about their daily business. I am appealing directly to anyone who witnessed the incident or has information that could help our inquiries to come forward. The force is hoping to speak to anyone who saw a person running in the area or those who have mobile phone footage. The scene remains cordoned off, with police forensic examinations expected to continue over the weekend. \\
    \hline
    \textbf{Reference:} A \textcolor{amber}{16-year-old} boy has died after he was stabbed in a busy \textcolor{amber}{Leeds street}, prompting a murder inquiry. \\
    \textbf{Cliff:}  A \textcolor{amber}{15-year-old} boy has died after being stabbed in \textcolor{amber}{Leeds}. \\ 
    \textbf{Frost (ECPP, Drop):} A teenager has been stabbed to death in a "busy area" in a busy street. \\
    
    \textbf{Finetuned (w/o $\lcal$):} A \textcolor{amber}{16-year-old} boy has died after being stabbed in a street in \textcolor{amber}{Leeds}. \\

    \textbf{SLiC-NLI (w/o $f_{len}$):} A teenage boy has been arrested after a teenager died following a stabbing in a busy area of West Yorkshire. \\
    
    \textbf{SLiC-NLI (with $f_{len}$):} A teenage boy has died after being allegedly stabbed in a busy street in West Yorkshire. \\

    \hline
    \end{tabular}
    } 
    \caption{An XSum inputs and various system generated summaries. The text spans in \textcolor{amber}{amber} are hallucinated.}
    \label{fig:xsum_predictions} 
    \vspace{-0.4cm}
\end{figure}

\paragraph{Final Results and Human Evaluations}

\begin{table}[th!]
    \centering
    \footnotesize
    {
        \begin{tabular}{c|ccc}
        {} &  quality &  factual &  length \\
        \hline
        Frost (ECPP, Drop)         &     3.18 &     .76 &   17.57 \\
        Cliff         &     3.10 &     .69 &   18.18 \\
        Finetuned (w/o $\lcal$)     &     2.96 &     .67 &   17.77 \\
        SLIC-NLI (w/o $f_{len}$)    &     \textbf{3.43} &     \textbf{.85} &   18.82 \\
        SLIC-NLI (with $f_{len}$) &     3.21 &     \textbf{.82} &   15.54 \\
        \hline
        \textit{Reference}        &     2.94 &     .60 &   2.65 \\
        \end{tabular}
    }
    \caption{Human Evaluation results on XSum dataset.}
    \label{tab:human_eval}
\end{table}

Table~\ref{tab:final_results} present our final results on the corresponding test sets. We conducted human evaluation of the generated summaries. Table~\ref{tab:human_eval} shows that SLiC-NLI improves consistency of the summaries from 67\% to 85\% and the average quality scores from 2.96 to 3.43. The results of the experiments on all other datasets are summarized in Tables~\ref{tab:human_eval_appendix} (Appendix ~\ref{sec:appendix_human_eval}). We also present summary lengths for comparison. The results show that calibration consistently improves the quality and factuality of all generated summaries. Humans consistently prefer the calibrated model over the non-calibrated model. See Figure~\ref{fig:xsum_predictions} where we demonstrate one of the examples that was given to the raters, both SLIC version are the only two models that produced non-hallucinated summaries.


\begin{table}[th!]
    \centering    
    \resizebox{7.7cm}{!}{
    \begin{tabular}{c|c|cccc}
    \hline
        $w$ & \textbf{Decoding} & \textbf{P (all)} & \textbf{S (all)}  & \textbf{P (top-1)} & \textbf{S (top-1)}\\ \hline
        && \multicolumn{4}{c}{$\cdot10^{-1}$} \\\hline
        w/o ~ &  \multirow{6}{*}{\STAB{\rotatebox[origin=c]{90}{Beam15}}} & 0.12 & 1.58 & 1.38 & 1.45 \\ 
        0.01 & & \textbf{3.05} & \textbf{3.12} & \textbf{2.83} & \textbf{2.94} \\
        0.003 & & 0.48 & 2.35 & 2.28 & 2.20 \\ 
        0.001 &  &0.28 & 2.18 & 2.14 & 1.99 \\ 
        0.0003 & & 0.14 & 1.85 & 1.70 & 1.71 \\ 
        0.0001 & & 0.14 & 1.74 & 1.60 & 1.62 \\ \hline
    \end{tabular}
    }
    \caption{Correlation between the log-probabilities of our model and NLI. We run inference with various decodings and compute the Pearson (\textbf{P}) and Spearman(\textbf{S}) correlations.  For the \textbf{beam} decoding we either used all the outputs or top-1. }
    \label{tab:corr}
\end{table}

\paragraph{Correlation with probabilities}

We study how the log-probability of the calibrated model correlates with NL (Table \ref{tab:corr}). For the beam search we either take the top-1 summary or the full beam outputs and compute the correlation across the whole datasets. The sentence log-probability as before computed as a sum of individual log-probabilities.



\section{Conclusions}

In this work we present \textit{\textbf{SLiC-NLI}} --- a new method for improving factuality of abstractive summarization models. The method calibrates the likelihood of the generative model with a consistency metric measured by NLI models. 

SLiC-NLI achieves state-of-the-art results on both human evaluation and automatic metrics while being simple, effective and straight-forward to implement. 
We show that SLiC-NLI  achieves a 18\% (from 67\% to 85\%) increase in consistency of the summaries according to humans and 31\%  (from  49\% to 80\%)  according to automatic metrics on \xsum dataset.

\footnotetext[1]{Metrics are computed using lowercase prediction and reference summaries available at \url{https://github.com/yixinL7/BRIO/}.}
\footnotetext[2]{We use our own implementation of SLiC so that these numbers don't match the ROUGE scores reported in the original paper}

We believe that our method has the potential to improve quality and factuality of generated text in a variety of applications. In future work, we plan to investigate the use of our method with other types of models, such as instruction-tuned models of size PALM-2~\cite{palm} and GPT-4~\cite{openai2023gpt4}. 
We hope that our work will contribute to the development of more reliable and accurate natural language generation systems.
\newpage


\section*{Limitations}

While SLiC-NLI is a powerful and simple method for improving consistency of summarization models, it is important to acknowledge its limitations. 
For example, we haven't explored the capabilities of the method beyond summarization tasks, and since the field of LLMs is moving fast in the direction of single unified models, it is important to make sure that our method works well with instruction-tuning techniques. 
Additionally, improved consistency does not always lead to a high performance in terms of other metrics. 
There are no guarantees that creativity and helpfulness of a model outputs will not be affected by improved consistency. Finding a natural balance and control of these aspects is one of the topics we would like to explore in the future work. 
Finally, even though our method is exceptionally good at increasing the consistency between summaries and the documents, it doesn't guarantee that other types of hallucinations that are not covered by NLI metric will not be generated.






\bibliography{anthology,custom}
\bibliographystyle{acl_natbib}

\newpage

\appendix


\section{Length regularisation Pareto frontier}
We compare how $\beta$ weight of length regularizer affects the other properties of the generated summaries, such as NLI, ROUGE2, Length, Coverage and Repetition. 
\label{sec:appendix_lr_pareto}
\begin{figure*}
\centering
    \includegraphics[width=2.1\columnwidth]{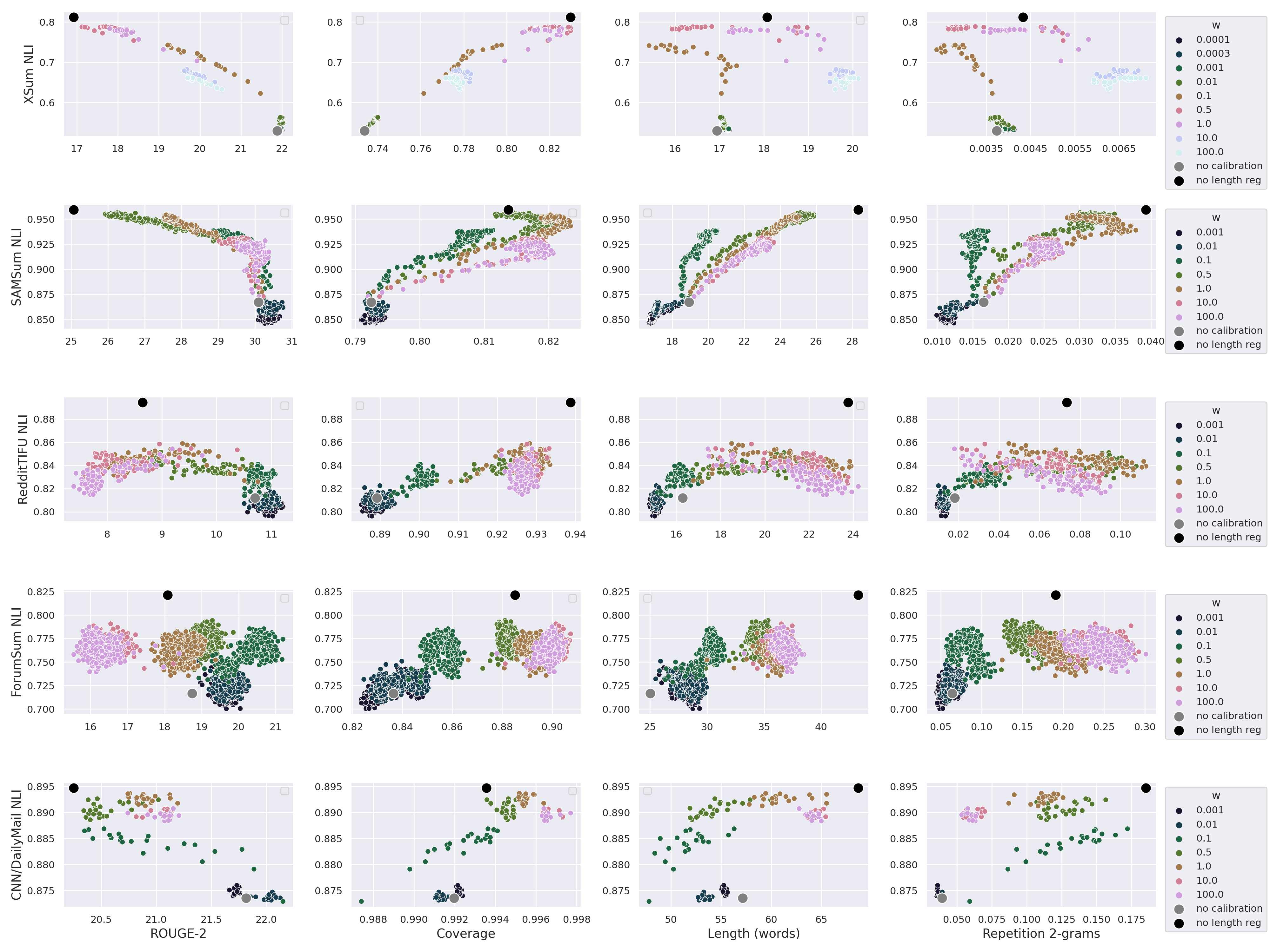}
\caption{Pareto frontier for the length regularization weight. We compare the affect of the different length regularizes on metrics such as NLI, ROUGE2, Coverage, Length and Repetition.}
\label{fig:app_length_reg}
\end{figure*}

\section{Length regularisation ablation}
See Figure~\ref{tab:len_cal_reg_app} for the ablation study of how different length regularizers affect the model length, repetition, NLI and ROUGE metrics.

\begin{table*}[t!]
    \centering
    \footnotesize
    {
    \begin{tabular}{cccccccc}
    \hline
        \textbf{Dataset} & \textbf{$\alpha$} &\textbf{NLI \%}  & \textbf{NLI gain \%} & \textbf{R1/R2/RL} & \textbf{Coverage \%} &
        \textbf{Length} & \textbf{Avg} \\
        \hline
                &                       100 &  89.07\% &    1.72\% &  43.96 / 21.15 / 30.72 &                  62.78 &                          99.6\% &               0.60 \\
                &                        10 &  89.07\% &    1.71\% &  43.88 / 21.10 / 30.60 &                  65.21 &                          99.7\% &               0.52 \\
                &                         1 &  89.37\% &    2.01\% &  43.14 / 20.76 / 30.36 &                  59.44 &                          99.5\% &               0.77 \\
                &                       0.5 &  89.26\% &    1.91\% &  42.42 / 20.46 / 30.15 &                  54.37 &                          99.5\% &               \textbf{0.83} \\
\textbf{\cnn}   &                       0.1 &  88.69\% &    1.33\% &  42.41 / 20.52 / 30.36 &                  56.30 &                          99.4\% &               0.69 \\
                &                      0.01 &  87.47\% &    0.11\% &  44.42 / 22.06 / 32.00 &                  53.27 &                          99.1\% &               0.40 \\
                &                     0.001 &  87.60\% &    0.24\% &  44.08 / 21.73 / 31.57 &                  55.26 &                          99.2\% &               0.43 \\
                &  SLIC-NLI (w/o $f_{len}$) &  89.47\% &    2.12\% &  42.41 / 20.25 / 29.78 &                  68.68 &                          99.4\% &               0.50 \\
                &   Finetuned (w/o $\lcal$) &  87.36\% &     0.00 &  44.29 / 21.82 / 31.62 &                  57.13 &                          99.2\% &               0.37 \\
         \hline
                  &                       100 &  78.76\% &    7.10\% &  35.86 / 15.96 / 28.52 &                  36.71 &                          90.0\% &               0.52 \\
                  &                        10 &  79.07\% &    7.41\% &  36.10 / 16.54 / 29.14 &                  36.52 &                          90.2\% &               0.54 \\
                  &                         1 &  78.27\% &    6.61\% &  37.65 / 18.34 / 30.94 &                  35.28 &                          89.5\% &               0.53 \\
                  &                       0.5 &  79.43\% &    7.77\% &  39.80 / 19.29 / 32.56 &                  35.35 &                          88.1\% &               0.59 \\
\textbf{\forumsum}&                       0.1 &  78.54\% &    6.89\% &  41.23 / 20.29 / 33.99 &                  30.52 &                          85.8\% &               \textbf{0.68} \\
                  &                      0.01 &  75.09\% &    3.43\% &  40.12 / 18.91 / 32.70 &                  25.92 &                          84.0\% &               0.64 \\
                  &                     0.001 &  74.40\% &    2.74\% &  40.21 / 19.26 / 32.86 &                  26.21 &                          83.8\% &               0.60 \\
                  &  SLIC-NLI (w/o $f_{len}$) &  82.13\% &   10.47\% &  38.51 / 18.08 / 31.16 &                  43.25 &                          88.5\% &               0.50 \\
                  &   Finetuned (w/o $\lcal$) &  71.66\% &     0.00 &  39.82 / 18.74 / 32.37 &                  25.03 &                          83.6\% &               0.50 \\
        \hline
                 &                       100 &  85.43\% &    4.22\% &   28.21 / 9.19 / 22.50 &                  17.42 &                          92.9\% &               \textbf{0.62} \\
                 &                        10 &  85.85\% &    4.64\% &   27.87 / 8.96 / 22.16 &                  19.40 &                          92.8\% &               0.53 \\
                 &                         1 &  85.89\% &    4.69\% &   27.68 / 9.38 / 22.59 &                  18.50 &                          92.8\% &               0.59 \\
                 &                       0.5 &  85.15\% &    3.95\% &   26.41 / 8.98 / 21.53 &                  19.67 &                          92.8\% &               0.48 \\
\textbf{\reddit} &                       0.1 &  84.11\% &    2.91\% &  29.82 / 10.76 / 24.69 &                  16.14 &                          90.0\% &               0.61 \\
                 &                      0.01 &  82.25\% &    1.05\% &  30.26 / 10.96 / 25.13 &                  15.09 &                          89.0\% &               0.49 \\
                 &                     0.001 &  81.59\% &    0.38\% &  30.39 / 10.79 / 24.96 &                  15.37 &                          89.1\% &               0.45 \\
                 &  SLIC-NLI (w/o $f_{len}$) &  89.43\% &    8.23\% &   27.28 / 8.65 / 21.60 &                  23.76 &                          93.9\% &               0.50 \\
                 &   Finetuned (w/o $\lcal$) &  81.21\% &     0.00 &  30.22 / 10.70 / 24.63 &                  16.28 &                          88.9\% &               0.43 \\
        \hline
                 &                   100 &  92.84\% &    6.10\% &  54.80 / 30.28 / 45.24 &                  22.88 &                          81.7\% &                   0.62 \\
                 &                        10 &  93.15\% &    6.41\% &  54.13 / 29.65 / 44.87 &                  22.98 &                          81.8\% &               0.63 \\
                 &                         1 &  95.41\% &    8.68\% &  52.16 / 27.59 / 42.92 &                  24.77 &                          82.0\% &               0.67 \\
                 &                       0.5 &  95.63\% &    8.90\% &  51.02 / 26.23 / 41.86 &                  25.13 &                          81.5\% &               0.66 \\
\textbf{\samsum} &                       0.1 &  93.89\% &    7.15\% &  53.47 / 29.15 / 44.76 &                  20.46 &                          81.0\% &               \textbf{0.80} \\
                 &                      0.01 &  86.90\% &    0.17\% &  54.60 / 30.10 / 45.84 &                  18.72 &                          79.2\% &               0.50 \\
                 &                     0.001 &  86.73\% &    0.00\% &  54.52 / 30.09 / 45.75 &                  18.93 &                          79.2\% &               0.49 \\
                 &  SLIC-NLI (w/o $f_{len}$) &  96.14\% &    9.41\% &  48.93 / 24.68 / 39.76 &                  29.08 &                          81.3\% &               0.50 \\
                 &   Finetuned (w/o $\lcal$) &  86.73\% &     0.00 &  54.52 / 30.09 / 45.75 &                  18.93 &                          79.2\% &               0.49 \\
        \hline
          &                       100 &  66.30\% &   13.28\% &  42.76 / 19.73 / 34.44 &                  20.02 &                          77.4\% &               0.24 \\
          &                        10 &  68.24\% &   15.22\% &  42.73 / 19.68 / 34.44 &                  19.62 &                          77.8\% &               0.32 \\
          &                         1 &  78.34\% &   25.32\% &  40.72 / 17.80 / 32.87 &                  18.54 &                          82.6\% &               0.62 \\
          &                       0.5 &  78.87\% &   25.85\% &  40.17 / 17.64 / 32.80 &                  16.82 &                          82.2\% &               \textbf{0.81} \\
          &                       0.1 &  74.28\% &   21.27\% &  41.36 / 19.22 / 34.11 &                  15.69 &                          79.8\% &               0.73 \\
          \textbf{\xsum} &                      0.01 &  56.35\% &    3.33\% &  44.82 / 21.96 / 37.02 &                  17.02 &                          74.0\% &               0.41 \\
          &                     0.001 &  53.62\% &    0.60\% &  44.89 / 21.99 / 37.06 &                  17.16 &                          73.3\% &               0.34 \\
          &                    0.0003 &  53.54\% &    0.53\% &  44.88 / 21.97 / 37.02 &                  17.20 &                          73.3\% &               0.33 \\
          &                    0.0001 &  53.39\% &    0.37\% &  44.86 / 21.93 / 37.00 &                  17.21 &                          73.3\% &               0.33 \\
          &  SLIC-NLI (w/o $f_{len}$) &  81.21\% &   28.19\% &  39.46 / 16.92 / 31.88 &                  18.07 &                          83.0\% &               0.73 \\
          &   Finetuned (w/o $\lcal$) &  53.02\% &    0.00\% &  44.73 / 21.88 / 36.94 &                  16.94 &                          73.4\% &               0.36 \\
        \
    \end{tabular}
    }
     \caption{
     The effect of various length regularizer weights on performance on all 5 datasets.} 
     \label{tab:len_cal_reg_app}
\end{table*}

\section{Experimental set up}

We run all experiments on T5 Large pre-trained checkpoints (770 million parameters),  using open-sourced T5X framework\footnote{https://github.com/google-research/t5x}.  For the infrastructure set up we used v3 TPU with 4 × 4 topology. Depending on the dataset the training can take from 5 hours (ForumSum) up to 7 days (CNN/DailyMail).

Reported metric results are collected from a single evaluation run on a test set, unless stated otherwise. 

For each dataset we first train a finetuned checkpoint where we swept the hyperparameters(checkpoint step, leaning rate, number of training steps) such as to achieve the top scores on the selected metrics. We used the validation set to choose the best finetuning checkpoint. Later at the calibration step we swept over various calibration loss weights $\alpha$ and for the length regularization results we chose the best result based on the sweep over the $\beta$ parameter. We used validation set again to pick the best checkpoint for the final results.

\section{Human Evals}
\label{sec:appendix_human_eval}

\begin{figure*}
\centering
    \includegraphics[width=2.1\columnwidth]{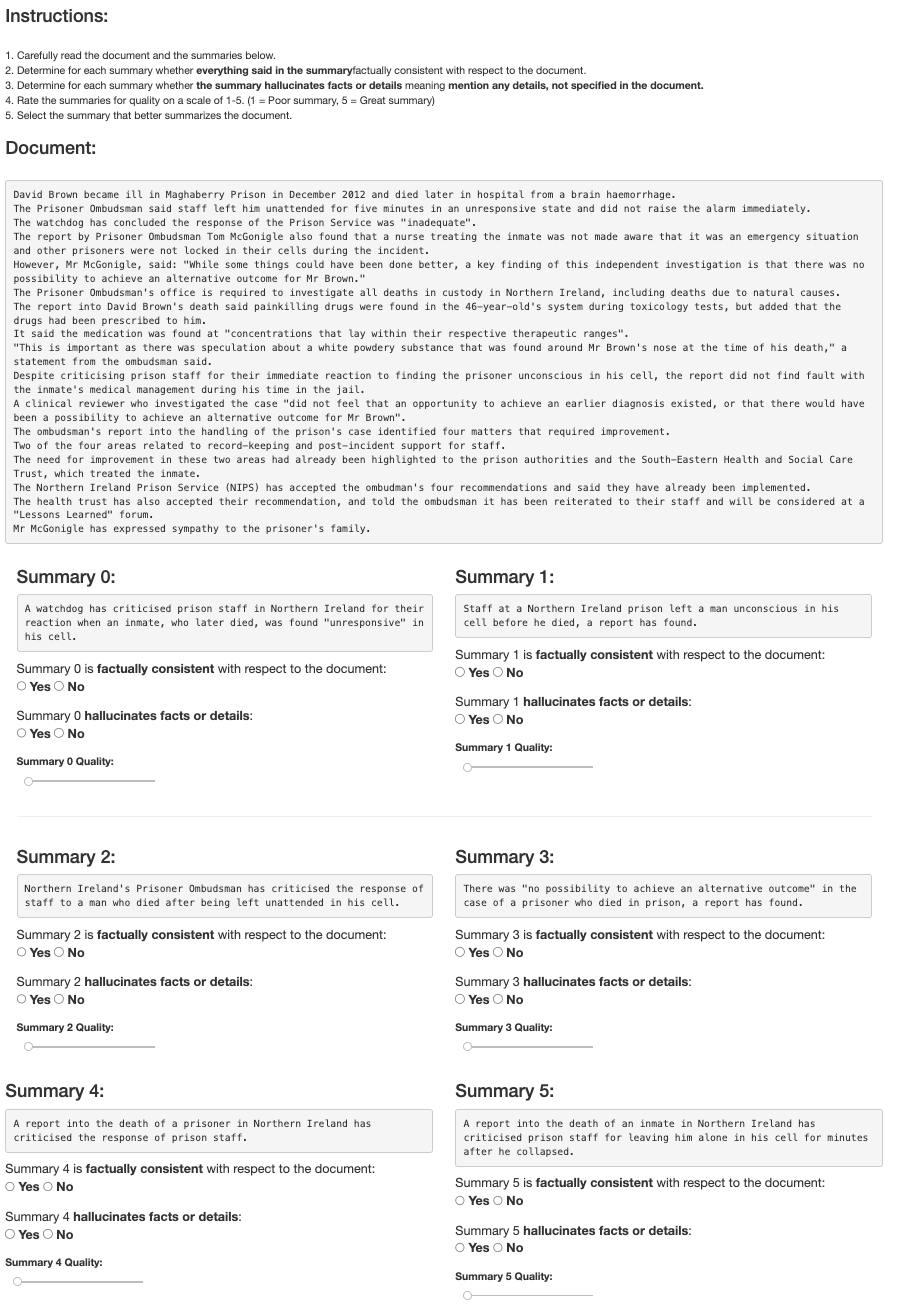}
\caption{An example of the task template that was offered to Amazon Mechanical Turk workers.}
\label{fig:human_eval_template}
\end{figure*}

Figure~\ref{fig:human_eval_template} presents our human evaluation template and instructions presented to our AMT workers. 
Table~\ref{tab:human_eval_appendix} shows our complete human evaluation results on all 5 datasets.

More examples of summaries before and after calibration can be found on Figure~\ref{fig:appendix_examples}.

\begin{figure}[t!]
    \centering
    \footnotesize
    \resizebox{0.99\columnwidth}{!}{
    \begin{tabular}{@{}p{8cm}@{}} \hline
    \textbf{Input: (SAMSum)} \\
    Raymond: Charlotte! Help! \\
    Charlotte: What's up bro?? \\
    Raymond: What do I want to eat, pizza or pasta? \\
    Charlotte: Hmm.. What kind of pizza and what kind of pasta? \\
    Raymond: So I have a regular cheese and pepperoni pizza and I was thinking some pesto pennes. \\
    Charlotte: Oo, those both sound good.  \\
    Raymond: That's not helpful. \\
    Charlotte: Have the pizza! \\
    Raymond: But pasta sounds so good.  \\
    Charlotte: Then have the pasta silly. \\ 
    Raymond: But the pizza sounds delicious too. \\
    Charlotte: Omg Raymond, make up your mind. \\
    Raymond: I can't! Please help me. \\
    Charlotte: Why not have both?\\
    Raymond: Well, that's just unreasonable no? \\
    Charlotte: How about this. I come over and we have both! \\
    Raymond: That could work. That way I would eat the same amount but of the two things I want to eat. \\
    Charlotte: Alright, so I'm going to head over in like 10 minutes. That sound good? \\
    Raymond: For sure. Oh, and bring wine! \\
    Charlotte: Yes sir. See you in 15.  \\
    \textbf{Before:} Raymond wants to eat pizza and pasta. Charlotte will come over in 10 minutes and they will have both. \textcolor{amber}{Raymond will bring wine.} \\
    \textbf{After:} Raymond wants to eat pizza or pasta. He has a regular cheese and pepperoni pizza and pesto pennes. Charlotte wants to come over in 10 minutes and they have both. Raymond wants Charlotte to bring wine. They'll see each other in 15. \\  
    \hline
    \textbf{Input: (SAMSum)} \\
    Amal: hey, did you see what Beyonce tweeted? \\
    Amir: haha. yeah, i did. isnt she wonderful? \\
    Amal: yeah, shes great. \\
    \textbf{Before:} Amal and Amir are \textcolor{amber}{laughing at} Beyonce's tweet. \\
    \textbf{After:} Amir saw what Beyonce tweeted. Amal thinks Beyonce is wonderful. \\  
    \hline
    \end{tabular}
    } 
    \caption{Inputs and system generated summaries before and after calibration for different datasets. The text spans in \textcolor{amber}{amber} are hallucinated.}
    \label{fig:appendix_examples} 
\end{figure}

\begin{table*}[t!]
    \centering
    \footnotesize
    {
    \begin{tabular}{cc|ccccc}
    \hline
            &        & \textbf{\samsum }& \textbf{\forumsum} & \textbf{\reddit} &  \textbf{\xsum} &  \textbf{\cnn} \\
    \hline
    \textbf{\em factual} 
            & Cliff &       ---               &    ---           &    ---           &  .69          &  \textbf{.99} \\
            & Frost (ECPP, Drop) &       ---               &    ---           &    ---           &  .76          &  \textbf{.99} \\
            & Finetuned (w/o $\lcal$)   &   \textbf{.88}   &    \textbf{.92} &    \textbf{.88} &  .67          &  \textbf{.98} \\
            & SLIC-NLI (w/o $f_{len}$) &   \textbf{.93} &    \textbf{.89} &    \textbf{.91} &  \textbf{.85} &  \textbf{.98} \\
            & SLIC-NLI (with $f_{len}$) &  \textbf{.91} &    \textbf{.94} &    \textbf{.89} &  \textbf{.82} &  \textbf{.98} \\
    \hline
            & Reference &   \textit{.94} &     \textit{.97}   &     \textit{.79} &  \textit{.60} &  \textit{.97} \\
    \hline
    \hline

    \textbf{\em quality} & Cliff &    ---    &     ---     &         ---         &  3.10 &           3.75 \\
            & Frost (ECPP, Drop) &     ---   &     ---     &       ---           &  3.18 &           3.60 \\
            & Finetuned (w/o $\lcal$)  &   3.41 &     3.46 &             3.19 &  2.96 &           3.62 \\
            & SLIC-NLI (w/o $f_{len}$) &   3.94 &     3.44 &             3.18 &  3.43 &           3.70 \\
            & SLIC-NLI (with $f_{len}$) &   3.54 &     3.48 &             3.11 &  3.21 &           3.78 \\
    \hline
            & Reference &   3.63 &     3.81 &             3.18 &  2.94 &           3.48 \\
    \hline
    \hline

    \textbf{\em length} & Cliff &   ---     &     ---     &       ---           & 18.18 &          59.85 \\
            & Frost (ECPP, Drop) &   ---     &      ---    &     ---             & 17.57 &          56.10 \\
            & Finetuned (w/o $\lcal$)   &  18.99 &    31.37 &            14.37 & 17.77 &          49.85 \\
            & SLIC-NLI (w/o $f_{len}$) &  31.17 &    37.05 &            18.53 & 18.82 &          66.56 \\
            & SLIC-NLI (with $f_{len}$) &  21.04 &    28.65 &            15.07 & 15.54 &          63.24 \\
    \hline
            & Reference &  20.47 &    34.32 &            19.77 & 20.65 &          53.86 \\
    \hline
    \end{tabular}
    }
    \caption{Human Evaluation results on all 5 datasets.}
    \label{tab:human_eval_appendix}
\end{table*}

\end{document}